\documentclass[10pt, conference]{IEEEtran}
\usepackage{graphicx}
\usepackage{hyperref}
\usepackage{xcolor}
\usepackage{multirow}
\usepackage{placeins}
\usepackage{amsmath}
\usepackage{lineno}
\usepackage{arydshln}
\usepackage{subfig}
\usepackage{amsfonts}
\usepackage{booktabs}
\usepackage{algorithm}
\usepackage{algorithmic}

\definecolor{maroon}{rgb}{0.5, 0, 0}

\newcommand{\be}{\begin{eqnarray}}
\newcommand{\ee}{\end{eqnarray}}
\newcommand{\bes}{\begin{eqnarray*}}
\newcommand{\ees}{\end{eqnarray*}}

\newcommand{\bigO}{\mathbb{O}}

\newcommand{\eg}{{\it e.g.}, }
\newcommand{\etal}{{\it et~al.}}

\begin{document}

\title{ACCEPT: Diagnostic Forecasting of Battery Degradation
Through Contrastive Learning}


\author{
\IEEEauthorblockN{James Sadler, Rizwaan Mohammed, Michael Castle, Kotub Uddin}
\IEEEauthorblockA{\textit{Envision Energy}, London, United Kingdom \\
Emails: \texttt{james.sadler@envision-energy.com}, \texttt{rizwaan.mohammed@envision-energy.com}, \\
\texttt{michael.castle@envision-energy.com}, \texttt{kotub.uddin@envision-energy.com}}
}

\maketitle

\begin{abstract}
Modeling lithium-ion battery (LIB) degradation offers significant cost savings and enhances the safety and reliability of electric vehicles (EVs) and battery energy storage systems (BESS).
Whilst data-driven methods have received great attention for forecasting degradation, they often demonstrate limited generalization ability and tend to underperform particularly in critical scenarios involving accelerated degradation, which are crucial to predict accurately. These methods also fail to elucidate the underlying causes of degradation. Alternatively, physical models provide a deeper understanding, but their complex parameters and inherent uncertainties limit their applicability in real-world settings. To this end, we propose a new model -- ACCEPT. Our novel framework uses contrastive learning to map the relationship between the underlying physical degradation parameters and observable operational quantities, combining the benefits of both approaches. Furthermore, due to the similarity of degradation paths between LIBs with the same chemistry, this model transfers non-trivially to most downstream tasks, allowing for zero-shot inference. Additionally, since categorical features can be included in the model, it can generalize to other LIB chemistries. This work establishes a foundational battery degradation model, providing reliable forecasts across a range of battery types and operating conditions.
\end{abstract}

\section{Introduction}
Lithium-ion batteries (LIBs) are a vital technology for advancing the transition from fossil fuels towards renewable energy solutions \cite{IPCC_2022_WGIII_Ch_6_SM}. Their high energy density, long cycle life, and steadily decreasing costs \cite{D0EE02681F} have spurred rapid adoption in both electric vehicles (EVs) and grid-scale battery energy storage systems (BESS). Modeling the inherent degradation of LIBs is challenging for several reasons. First, the rapid development of LIB technology means that operational or experimental datasets with batteries near their end-of-life are difficult to obtain. Second, the degradation is driven by several internal non-linear chemical processes, and depends strongly on the operating conditions. In particular, degradation, sometimes referred to as State of Health (SoH), is affected by the number of operating cycles, temperature, charge/discharge rate, and depth of discharge \cite{review_factors}. Battery degradation can be accurately parameterized by the combination of loss of Li$^+$ inventory (LLI) and loss of active material (LAM) \cite{BIRKL2017373,xu2023high}.

Methods for modeling degradation can be broadly divided into two categories -- physics-based and data-driven techniques. Physics-based techniques attempt to model the underlying physical and chemical mechanisms that cause degradation, such as lithium plating \cite{SAHU2023106516} and solid-electrolyte interface (SEI) growth \cite{SEI_model}. On the other hand, data driven methods predominantly use operational characteristics to predict the future capacity.

Data-driven methods include recursive algorithms such as Kalman filters \cite{MU2024110221} and Sequential Monte Carlo methods \cite{en12142784}. Whilst these techniques can yield useful predictions, they are model dependent and struggle to handle measurement noise and inaccuracies effectively. Consequently, there has been a growing shift toward time-series machine learning models for forecasting lithium-ion battery (LIB) degradation, including recurrent neural networks (RNNs), long short-term memory (LSTM) networks, and convolutional neural networks (CNNs) \cite{su141911865, 9137406}.

Although deep-learning models have achieved some success in forecasting LIB degradation, most studies primarily focus on estimating remaining useful life (RUL) or capacity curves. These approaches face two significant limitations:  First, they generalize poorly to conditions not seen in the training set and often fail to predict knee-points, particularly in cases with unusual degradation paths \cite{Attia_2022}. Second, they make no attempt to diagnose the degradation by quantifying the underlying LLI and LAM. This work aims to address both of these challenges.

On the other hand, generating data purely from pre-parameterized physics-based models (either through direct simulation or training data-driven models on synthetic datasets) is often insufficient for accurate degradation prediction. Real-world factors such as manufacturing defects, subtle variations in operating conditions, or inaccuracies in model parameters can lead to large discrepancies between simulated and actual battery performance \cite{miguel2021review}. Whilst parametrizing the physics-based models can mitigate this, the process is computationally expensive, has strict data requirements, and depends heavily on knowledge of LIB design. Additionally, the accuracy of physics-based models relies on a detailed understanding of the degradation mechanisms for the specific LIBs in question -- an ongoing area of research with no one-size-fits-all solution \cite{chen2016overview, o2022lithium, reniers2019review, birkl2017degradation, edge2021lithium}. Therefore, in many real-world scenarios where data is incomplete and real-time solutions are required, purely physics-based approaches can be impractical.

An alternative approach is offered by Dubarry \etal, who employ a modified equivalent-circuit model (ECM) to emulate LIB performance under various states of degradation \cite{dubarry2012synthesize}. The authors propose a broad set of potential degradation behaviours that the ECM can simulate, thus reproducing voltage curves and (dis)charging characteristics for a wide variety of degradation paths. By comparing these simulated curves with experimental data, Dubarry \etal~demonstrate that it is possible to identify the state of degradation in LIBs and infer their physical state, including LAM and LLI. However, this method doesn't allow accurate degradation forecasting, as different degradation pathways are often similar in the early stage, so extrapolating a simulated curve introduces prohibitively large uncertainties.

In this work, we introduce ACCEPT\footnote{Adaptive Contrastive Capacity Estimation Pre-Training}, a novel model that forecasts degradation by combining the strengths of both data-driven and physics-based approaches. The model structure is inspired by OpenAI's CLIP \cite{CLIP}, which demonstrated the power of contrastive learning for zero-shot prediction. ACCEPT learns to predict the most probable future degradation path from a large range of simulated curves, by using combination of historical capacity sequences and operational features, including temperature, current, and voltage.

This approach offers a significant advantage: by incorporating a physics-based battery model, we can quickly and cheaply simulate a wide range of degradation paths without the need for extensive experimentation. These simulated curves can be matched to capacity curves from real, operational data of batteries in open-source datasets to create a labeled dataset. Thus, information from both the operational and simulated data is leveraged to accurately forecast the degradation path.

Using the physics-based model, each simulated data point can be associated with a combination of LLI and LAM. The operational data and the simulated data are encoded separately using time-series models. These encoded representations are then projected into a shared embedding space, enabling the model to learn complex relationships between observed and simulated degradation patterns. During inference, an LIB's operational features are encoded and the closest matching simulated degradation curve is retrieved. This simulated curve is taken as the prediction for its future degradation.

To embed the operational data, a modified Temporal Fusion Transformer (TFT) \cite{lim2021temporal} is used. This allows battery metadata, such as the chemistry of the LIB and the initial capacity, to be included in the model, allowing the model to be generalized across different LIB types. To embed the simulated degradation curves, a CNN-based architecture is used. This retrieval-based method allows the quantification of degradation modes, which is not feasible in existing data-driven methods. 

This approach allows for the degradation pathway to be estimated from as few as 100 cycles. We show that this generalizes better than existing approaches to unknown operational scenarios, helping to mitigate the data availability problem encountered in battery research. Additionally, if the parameters of an LIB are not known and therefore curves cannot be generated using physics-based models, users can still obtain accurate results using curves generated for LIBs with similar chemistries.

Sec.~\ref{sec:approach} outlines the method for generating the simulated curves and then describes the model in detail. The model training and results are shown in Sec.~\ref{sec:experiments}.

\section{Proposed Approach} \label{sec:approach}
The proposed approach is to train a model to match a set of operational data associated with a specific degradation pathway with the corresponding simulated scenario. This allows both diagnosis of historic degradation and forecasting of the future capacity fade. The operational and simulated data are embedded separately to perform the comparison. Features such as temperature, voltage, and current are included in the operational embedding so that the model can generalize to unseen scenarios. Before training, a set of simulated curves is generated from a simple set of degradation equations, as described in Sec.~\ref{sec:generating_curves}. The model architecture is shown in Sec.~\ref{sec:architecture}.

\begin{figure*}
    \centering
    \includegraphics[width=1\linewidth]{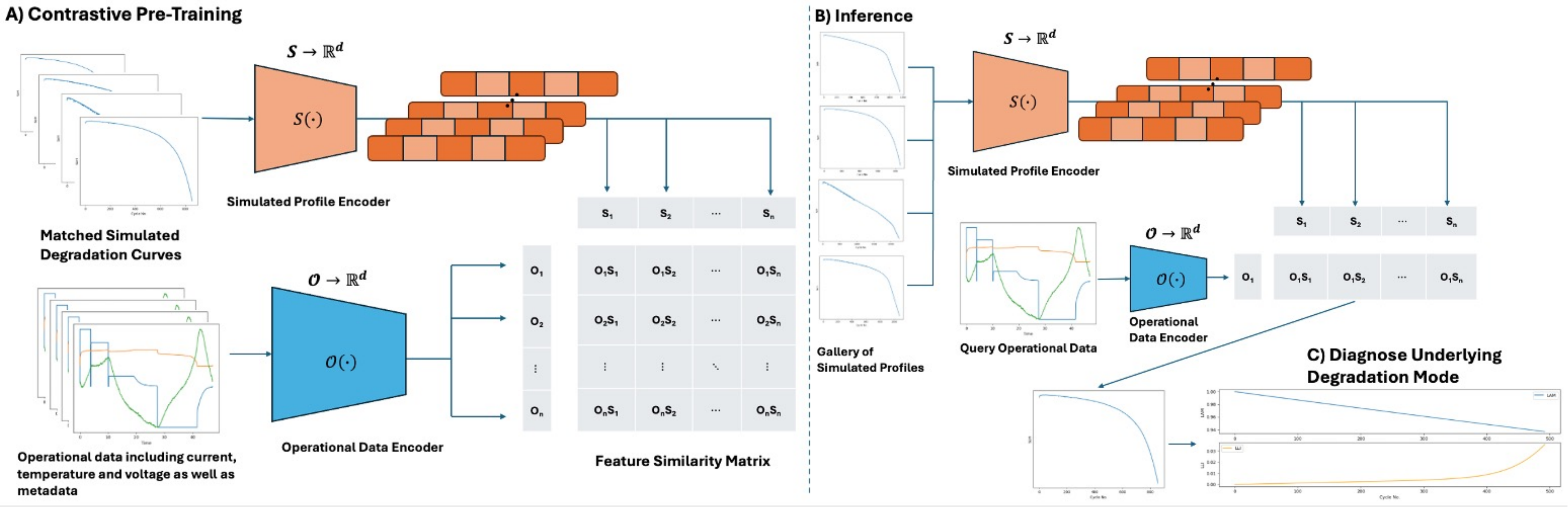}
    \caption{Summary of approach. Whilst standard data-driven techniques directly train a time-series model to predict the time-series, this model trains an operational data (current, temperature, voltage, static meta data inc. LIB chemistry and initial capacity) encoder and a simulated curve encoder. Curves are matched to their operational data to enable a zero-shot degradation prediction.}
    \label{fig:model_arc}
\end{figure*}

\subsection{Generating Simulated Degradation Curves} \label{sec:generating_curves}
The choice of degradation model is not fundamental to ACCEPT. In principle, any method capable of generating simulated degradation curves that sufficiently describe the full range of operational curves under study can be used. Below, we present a simple dimensionless set of equations as a proof of concept, but acknowledge that other methods, such as \cite{DUBARRY2020228806}, may be more suitable for real-world applications.

A mathematical model, in the form of a system of ordinary differential equations describing capacity fade as a combination of LLI and LAM, is proposed. This is built on the idea of \cite{dubarry2012synthesize, mohtat2019towards}, which proposes that capacity fade can be described as a combination of LLI and LAM. This model is built on simplified physics as the product of remaining active material and lithium inventory. The LIB's active material is assumed to degrade exponentially over time, while lithium lost is described as a combination of SEI growth and lithium plating. SEI growth is assumed to increase linearly, whereas for plating a time delay is included in order to simulate that this mode of degradation may not begin at start of life. This former mechanism aims to emulate the idea that plating often occurs only once local voltages within the LIB are large enough, an effect which occurs only during extreme operating conditions \eg high C-rates, low temperatures or unsafe voltage ranges, and once other degradation mechanisms have accumulated to cause such effects, \eg low porosity due to SEI growth or high effective C-rates due to reduced capacity \cite{chen2016overview, o2022lithium, reniers2019review, birkl2017degradation, edge2021lithium, lin2021lithium}.

During the inference process, any number of positive samples from one to infinity (constrained by available computing power) can be fed into the simulated profile encoder. The model will then return the similarity score of the operational data to each of these samples. This allows for the creation of custom simulated curves that can be used for specific use-cases, or curves from pre-existing data. Two curves can be compared to see how likely it is a LIB's trajectory will follow each of them (linear projection vs expected knee), or a large number can be used for accurate curve estimation. 

\paragraph{Degradation Equations\label{sec_degradation_model}}

The proposed dimensionless model that describes the decline of LIB capacity due to SEI growth and lithium plating in $t\in[0,\infty)$ is given by (\ref{eqn_LAM})-(\ref{eqn_param_tanh}). Here, $t$ can be defined as both time and cycle number, which are equivalent under transformations to the model parameters.
LAM degradation is modeled by,
\be \label{eqn_LAM}
\dot{M} = -k M,
\ee
where $M$ is the material-capacity of the LIB, with initial condition $M(0)=1$, and $k$ is the rate of material degradation. LLI degradation is modeled by,
\be \label{eqn_LLI}
\dot{S} = a, \quad \dot{P} = \begin{cases}0 & t\leq t_p,\\ 0.5 b \left(1+\tanh\left(c\left(t-t_p\right)\right)\right) & t>t_p,\end{cases}
\ee
where $S$ and $P$ are the lithium-loss due to SEI and plating, respectively, which have initial conditions $S(0)=P(0)=0$, $a$ and $b$ are growth rates of SEI and plating, respectively, $c$ determines the sharpness of the knee, and $t_p$ is the point in time which plating begins. Capacity is modeled by,
\be
L = S + P, \qquad C = (1-L)M,
\ee
where $L$ is the total loss of lithium inventory and $C$ is the LIB capacity. $L\in [0,1]$, where $L=1$ corresponds to the complete loss of lithium inventory. The SEI and plating growth rates, $a$ and $b$, are defined as functions of time by,
\be \label{eqn_param_tanh}
\begin{aligned}
a(t) &= 0.5 a_0 \left(1+\tanh\left(100\left(1-L\right)\right)\right), \qquad \\
b(t) &= 0.5 b_0 \left(1+\tanh\left(100\left(1-L\right)\right)\right),
\end{aligned}
\ee
where $a_0$ and $b_0$ are the typical rate parameters for $a$ and $b$, respectively.
In equations (\ref{eqn_LLI}) and (\ref{eqn_param_tanh}), the hyperbolic tangent function, $\tanh$, is used as a way to continuously activate or deactivate processes as the battery state evolves. In equation (\ref{eqn_LLI}), plating does not begin until $t\approx t_p$, which simulates a knee in the capacity curve, where the sharpness of the knee is determined by $c$. In equations (\ref{eqn_param_tanh}) LLI growth is switched off as $L\to1$, accounting for scenarios where all lithium inventory is lost.

\paragraph{Numerical Simulation}

The system of ordinary differential equations (\ref{eqn_LAM})-(\ref{eqn_param_tanh}) are solved numerically using the classic fourth-order Runge-Kutta method, RK4 \cite{griffiths2010numerical}, with a step size of $h = 0.01$. An early stop criterion is applied when $C<0.7$ (70\% SoH), since the experimental data does not include degradation beyond this point. The total accumulated error of the RK4 method is $\bigO(h^4)$ and hence, given the method's stability, the numerical error is negligible.

Each curve in the operational dataset is parametrized by finding optimal values for the parameters $k$, $a_0$, $b_0$, and $t_p$. This is achieved using Python's SciPy \cite{virtanen2020scipy} \texttt{minimize} function which minimizes the L2-norm (approximated by the midpoint rule) between the data and simulation.

\begin{figure}
    \centering
    \includegraphics[width=0.5\textwidth]{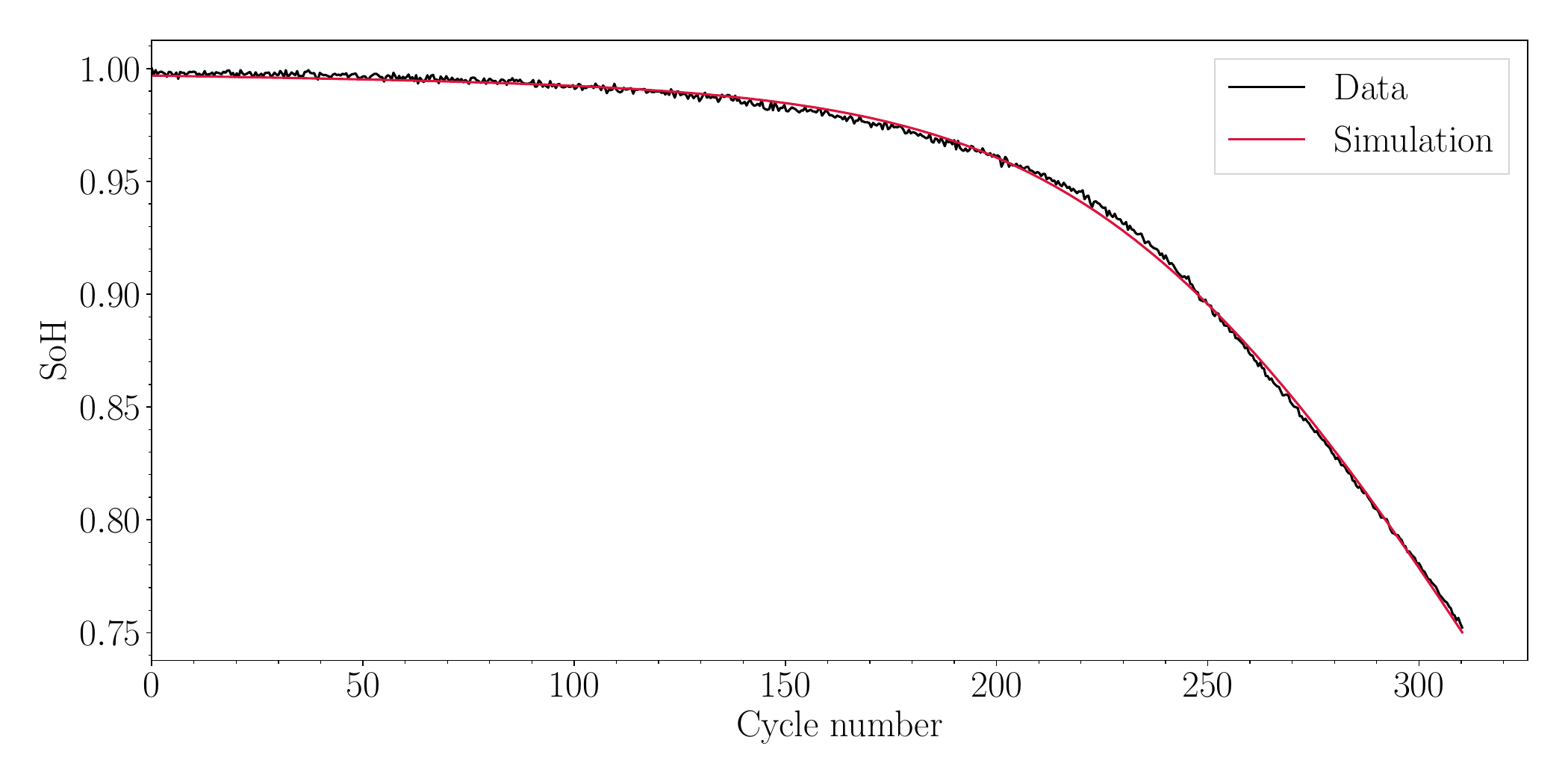}
    \caption{Example of an operational degradation curve with the corresponding matched simulated curve.}
    \label{fig:example_fit}
\end{figure}

To create a comprehensive training set, simulations are generated by solving the model for a large set of parameter values sampled uniformly within the ranges identified from the parameterized data curves. Fig.~\ref{fig:example_fit} shows an example of an operational curve and the best matching simulated curve.

\subsection{Model Architecture} \label{sec:architecture}
The architecture of ACCEPT is shown in Fig.~\ref{fig:model_arc}, and a description follows:

Let $\mathbb{S}$ and $\mathbb{O}$ represent the vector spaces for the simulated and operational data, respectively, where $\mathbb{S}$ consists solely of the capacity (degradation) curves, while $\mathbb{O}$ includes both the degradation curves and associated operational features such as temperature, voltage, and current. The architecture consists of two distinct encoders: (i) a simulation encoder $E_s: \mathbb{S} \rightarrow \mathbb{R}^{d}$, and; (ii) an operational encoder $E_o: \mathbb{O} \rightarrow \mathbb{R}^{d}$, where $d$ is the dimension of the shared latent space ($d=512$ is used in this case).

\begin{figure}[ht]
    \centering
    \includegraphics[width=\linewidth]{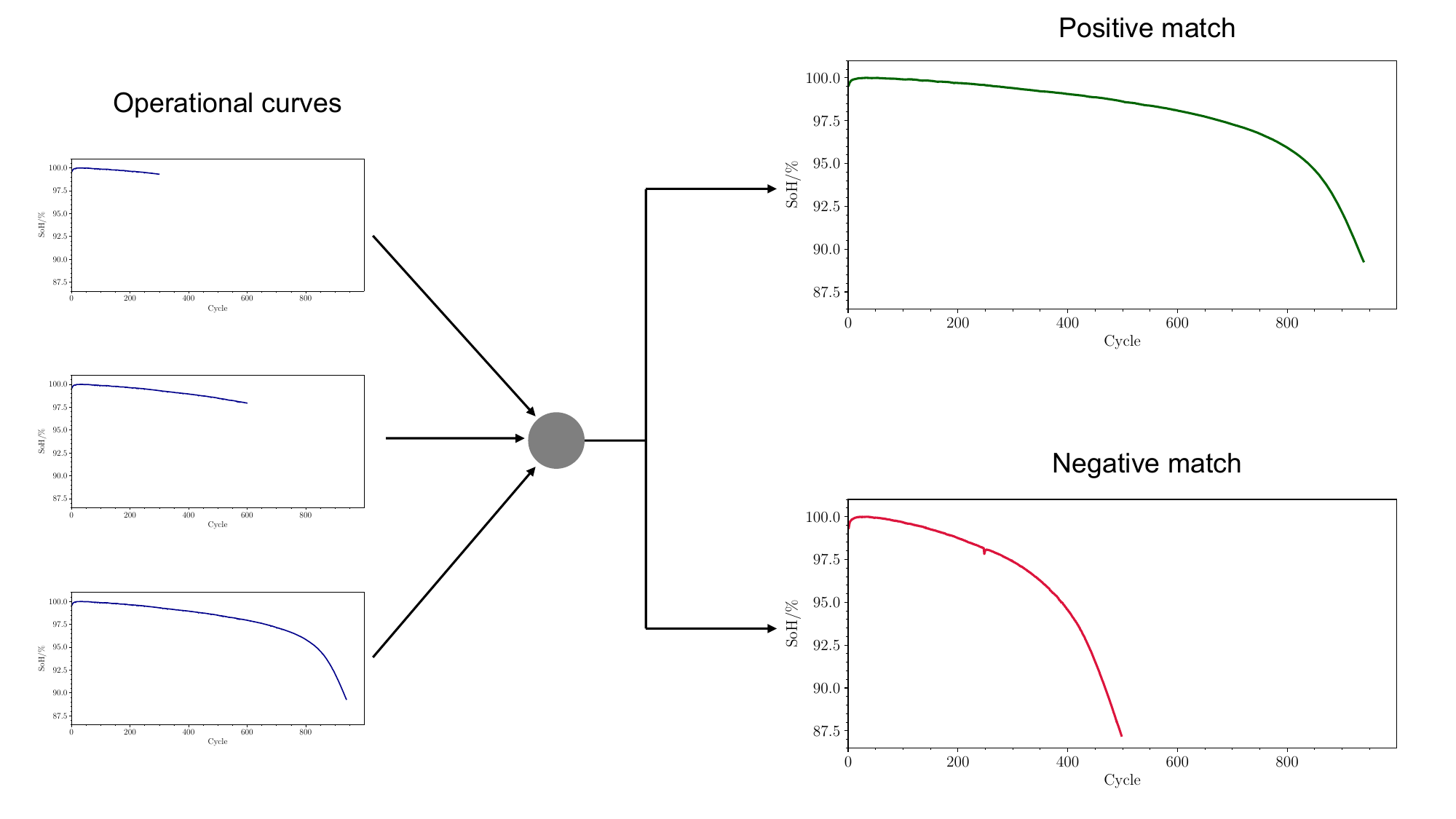}
    \caption{On the left of the image, a degradation curve from the same battery at different points in its lifecycle is shown. In the training process, each of these curves would form a positive and negative pair with the simulated curves on the upper right and lower right, respectively.}
    \label{fig:curves}
\end{figure}

The simulation encoder, $E_s$, employs a CNN architecture:
$$z_s = \text{Pool}(\text{CNN}(s; \theta_s)),$$
where CNN consists of $L$ layers of 1D convolutions with ReLU activations. Each layer $l$ applies:
$$x^{(l)} = \text{ReLU}(\text{Conv1D}(x^{(l-1)})) \in \mathbb{R}^{C_l \times T_l},$$
where $C_l$ represents the number of channels in layer $l$, and $T_l$ represents the temporal length of the sequence at that layer.
This model showed a good trade-off between extracting long-term trends and computational efficiency, although it is possible to replace this block with other architectures well suited for time-series processing.
\paragraph{Operational Encoding}
We adopt the TFT \cite{lim2021temporal} as the basis of the operational encoder due to its state-of-the-art performance in time-series forecasting. The TFT effectively captures temporal dependencies at multiple scales and includes specialized components such as the Variable Selection Network for identifying the most relevant features at each time step. Moreover, it naturally integrates static real-valued and categorical data, enabling us to incorporate crucial information like LIB type and chemistry. This flexibility is vital for the zero-shot setting, as we aim to learn general representations that transfer effectively to a wide range of battery configurations and operational conditions. In theory, any model that captures this information could be used as the operational encoder. When passing the data to the operational encoder, multiple input sequences of varying lengths are created from each operational curve, as shown in Fig.~\ref{fig:curves}. This enhances the model's ability to make predictions at different points in the LIB lifecycle.

The operational encoder $E_o$ modifies the TFT by replacing the forecasting head with a dense embedding layer:
$$h = \text{TFT}_{\text{base}}(o; \theta_o),$$
$$z_o = W_e h + b_e,$$
where:
\begin{itemize}
\item $\text{TFT}_{\text{base}}$ is the standard TFT encoder-decoder architecture until the final forecasting layer
\item $h \in \mathbb{R}^{d_h}$ is the final hidden state of the TFT
\item $W_e \in \mathbb{R}^{d \times d_h}$ and $b_e \in \mathbb{R}^{d}$ are learnable parameters of the embedding layer
\item $z_o \in \mathbb{R}^{d}$ is the final operational embedding
\end{itemize}

Specifically, $\text{TFT}_{\text{base}}$ processes the input through:
$$v_t = \text{VSN}(o_t),$$
$$c_t = \text{MHA}({v_1,...,v_T}),$$
$$h = \text{Pool}({c_1,...,c_T}),$$
where VSN is the variable selection network and MHA is multi-head temporal attention. Unlike the original TFT, we omit the quantile forecast layer and instead map the pooled representations directly to the embedding space.

\paragraph{Contrastive Learning}
Contrastive learning is a self-supervised learning technique that has gained significant attention in machine learning, especially in the domain of computer vision and natural language processing. It learns useful representations by comparing pairs of examples and pulling similar ones together while pushing dissimilar ones apart in a high-dimensional latent space \cite{chen2020simple}.

For a batch of paired samples ${(s_i, o_i)}_{i=1}^N$, the contrastive loss function can be constructed as:
\begin{equation} \label{eq:contrastive_loss}
\mathcal{L} = -\sum_i \log\frac{\exp(\text{sim}(p_{s_i}, p_{o_i})/\tau)}{\sum_j \exp(\text{sim}(p_{s_i}, p_{o_j})/\tau)} \;,
\end{equation}
where the normalized embeddings are:
\begin{equation}
\begin{aligned}
    p_{s_i} &= \displaystyle\frac{E_s(s_i)}{\bigl\|E_s(s_i)\bigr\|_2} \;, \\
    p_{o_i} &= \displaystyle\frac{E_o(o_i)}{\bigl\|E_o(o_i)\bigr\|_2} \;,
\end{aligned}
\end{equation}
$\text{sim}(a, b) = a^\top b$ is cosine similarity, and $\tau$ is a temperature parameter. $j$ indexes all positive and negative matches in the batch. In equation (\ref{eq:contrastive_loss}), $\tau$ is a learnable parameter which specifies the size of the penalty for negative matches and is determined as part of the training process.

Additionally, we employ a queue of negative pairs to enhance the in-batch negatives and diversify the examples learned by the model. Let $\mathbf{Q}$ be a queue that stores all the additional negative embeddings $\{p_{o_i}^{-}\}_{i=1}^M$. During training, for each batch, $K = 2048$ random elements are selected from $\mathbf{Q}$. For each positive pair $(p_{s_i}, p_{o_i})$ in a mini-batch, the denominator in Eq.~\eqref{eq:contrastive_loss} is replaced by this term:
\begin{equation}
\begin{aligned}
\alpha(p_{s_i}, p_{o_j}, \tau)
&= \sum_j \exp \bigl(\mathrm{sim}(p_{s_i}, p_{o_j})/\tau\bigr) \\
&\quad + \sum_{k=1}^{K} 
   \exp \bigl(\mathrm{sim}(p_{s_i}, p_{o_k}^{-})/\tau\bigr)\:,
\end{aligned}
\end{equation}
where $p_{o_k}^{-}$ represents a negative match from the queue. The full loss function becomes:
\begin{equation}
    \mathcal{L} = -\sum_i \log\frac{\exp(\text{sim}(p_{s_i}, p_{o_i})/\tau)}{\alpha(p_{s_i}, p_{o_j}, \tau)} \;.
\end{equation}
This design ensures that each sample is contrasted against a large and continually refreshed set of negatives, improving the robustness of the learned embeddings. The queue is dynamic, a different sample of negatives is randomly selected for each batch.

\subsection{Zero-shot}

When creating data-driven models for the estimation of battery degradation, a big problem faced by researchers is the poor transferability to downstream tasks. This is due to the inability of most data-driven models to generalize to unseen operating conditions. Through matching operational data to degradation curves using embedding models, ACCEPT is better able to handle unforeseen circumstances. Zero-shot learning refers to the ability of a model to generalize to new classes without being specifically trained on them \cite{xian2018zero}. Due to the large testing times of Li-ion batteries under development, as well as the importance of early detection of LIBs set to experience accelerated degradation, models with this capability are of great interest to the industry. 

\begin{algorithm}[h!]
   \caption{ACCEPT Training}
   \label{algorithm}
\begin{algorithmic}
\STATE \textbf{Input:} 
\STATE \vspace{0.25em}
   \begin{itemize}
     \item Training set $\mathcal{D} = \{(s_i, o_i)\}_{i=1}^{N}$ 
     \item Batch size $B$
     \item Penalty $\tau$
     \item Queue size $M$ (total number of additional negative embeddings)
     \item Encoders $E_s, E_o$ with parameters $\theta_s, \theta_o$
     \item Optimizer $\mathrm{ADAM}(\theta_s, \theta_o)$
   \end{itemize}
\STATE \vspace{-0.25em}
\STATE \textbf{Output:} Trained ACCEPT model parameters $(\theta_s, \theta_o)$
\vskip 1pt
\STATE Initialize the queue $\mathbf{Q}$ with precomputed negative embeddings; $\mathbf{Q} \gets \{p_{o_k}^{-}\}_{k=1}^{M}$
\REPEAT 
     \STATE Sample a mini-batch $\{(s_i, o_i)\}_{i=1}^B$ from $\mathcal{D}$
     \FOR{$i=1$ {\bfseries to} $B$}
       \STATE $z_{s_i} \gets E_s(s_i; \theta_s)$ \quad // 1D CNN encoder
       \STATE $z_{o_i} \gets E_o(o_i; \theta_o)$ \quad // TFT encoder
       \STATE $p_{s_i} \gets \mathrm{Normalize}(z_{s_i})$
       \STATE $p_{o_i} \gets \mathrm{Normalize}(z_{o_i})$
     \ENDFOR\
\STATE \vspace{-1em}
\STATE \textbf{Draw random sample of $K$ negatives from $\mathbf{Q}$:}
\STATE $\mathbf{Q}_{\mathrm{sample}} \gets \mathrm{RandomSample}\bigl(\mathbf{Q}, K\bigr)$
\STATE $\mathcal{L} \gets 0$
\FOR{$i=1$ {\bfseries to} $B$}
      \STATE \vspace{-3em}
       \STATE \small \begin{align*}
              \alpha_i(p_{s_i}, p_{o_j}, \tau)
                &= \sum_{j=1}^B 
                   \exp \bigl(\mathrm{sim}(p_{s_i}, p_{o_j})/\tau\bigr) \\
                &\quad + \sum_{k \in \mathbf{Q}_{\mathrm{sample}}}
                   \exp \bigl(\mathrm{sim}(p_{s_i}, p_{o_k}^{-})/\tau\bigr).
            \end{align*} \normalsize
        \STATE \vspace{-1em}
       \STATE \(\mathcal{L}_i
            = -\log \Biggl(
            \displaystyle\frac{\exp\bigl(\mathrm{sim}(p_{s_i}, p_{o_i})/\tau\bigr)}
            {\alpha_i(p_{s_i}, p_{o_j}, \tau)} \Biggr)
            \)
       \STATE $\mathcal{L} \gets \mathcal{L} + \mathcal{L}_i$
     \ENDFOR
 \STATE \textbf{Update parameters using optimizer:}
 \STATE \vspace{-0.5em}
     \[
       (\theta_s, \theta_o) \gets 
       \mathrm{ADAM}(\theta_s, \theta_o, \nabla \mathcal{L})
     \]
\UNTIL{convergence}
\STATE \textbf{return} $(\theta_s, \theta_o)$
\end{algorithmic}
\end{algorithm}

\section{Experiments} \label{sec:experiments}
\subsection{Dataset}
Severson et al. \cite{severson2019data} generated a comprehensive dataset consisting of 124 lithium iron phosphate/graphite LIBs that were cycled under fast-charging conditions. The experiments were stopped when the batteries reached end-of-life (EOL) criteria. EOL cycle number ranged from 150 to 2300. For zero-shot estimation, as described in Sec.~\ref{sec:zero_shot}, we used a separate dataset of LIBs with end-of-life (EOL) cycle numbers ranging from 450 to 1200, which was not included in the training process \cite{attia2020closed}.

\subsection{Training Procedure}
A pseudocode outline of the model training procedure is shown in Algorithm~\ref{algorithm}. ACCEPT was trained using an Adam optimizer \cite{ADAM} and a learning rate of $1\times 10^{-5}$. The model was trained until convergence, which typically happened around the 7th epoch. The model was relatively quick to train, taking around two hours on a single Nvidia A100 GPU. Eight LIBs were reserved from the total dataset as validation LIBs during the training process. 16 LIBs were used as the test and evaluation set.

\subsection{Accuracy Comparison with State-of-the-Art Results}

\begin{figure}[ht]
    \centering
    \includegraphics[width=0.5\textwidth]{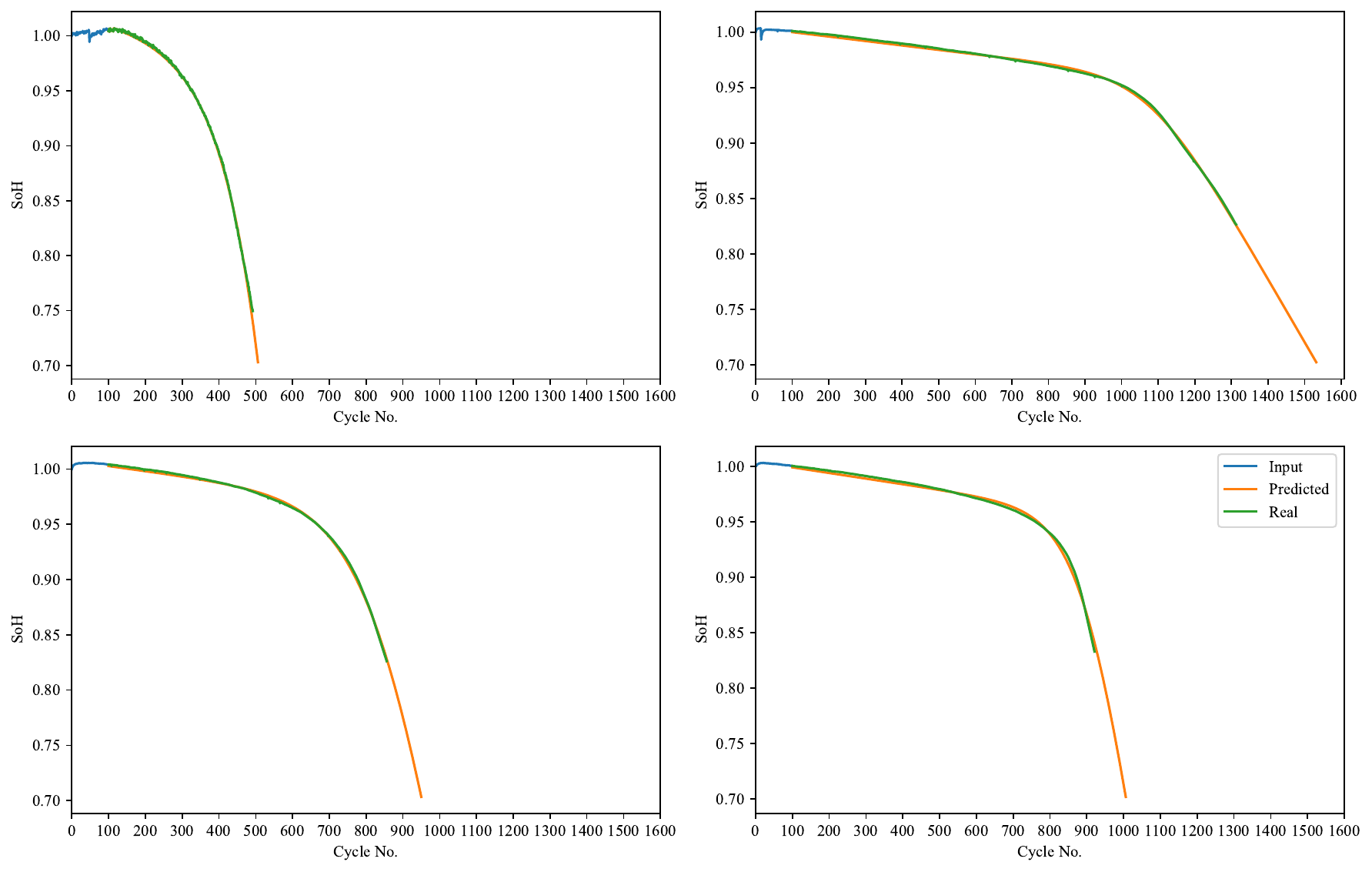}
    \caption{Future capacity degradation curves as predicted by the model against test LIBs no. 1-4 from 100 cycles of input data. The model has the added benefit against current techniques that it is not limited to a certain output (context) length.} 
    \label{fig:results}
\end{figure}

The model was compared to a variety of conventional data-driven techniques for estimation of future degradation. It was shown that by using as little as 100 cycles of input data the model was capable of producing accurate degradation curves. The prediction for several test cells is shown in Fig~\ref{fig:results}. An additional benefit of this approach is that we are not limited by the output sequence length of the model, meaning that we can return a degradation pathway of any length, rather than just the model output dimension $n$, as with other data-driven techniques. 

In Tab.~\ref{tab:results} it can be seen that ACCEPT achieves state-of-the-art results compared to existing methods for accuracy. 

\begin{table}[ht]
\begin{center}
\begin{sc}
\caption{Comparison of state-of-the-art results for degradation models, where proposed model uses inputs from 100 cycles. Results from the proposed method are averaged across all 16 test LIBs. The results show our method is the best-performing across all evaluation metrics.}
    \begin{tabular}{lccc}
    \toprule
   Method & MSE & MAE & MAPE\\
   \midrule
   Elman-LSTM \cite{li2019remaining} & 0.0054 & 0.0525 & 0.0902 \\
   TCN \cite{zhou2020state} & 0.0090 & 0.0522 & 0.0944 \\
   FPC \cite{mittal2023two} & 0.0035 & 0.0454 & 0.0882 \\
   ST-MAN \cite{suh2024remaining} &  0.0014 & 0.0275 & 0.0494 \\
   \midrule
   \textbf{Proposed}  & \textbf{0.0005} & \textbf{0.0082} & \textbf{0.0088}\\
\bottomrule
\end{tabular}

    \label{tab:results}
\end{sc}
\end{center}
\end{table}

\subsection{Quantifying Degradation Modes}
\begin{figure}[ht]
    \centering
    \includegraphics[width=0.5\textwidth]{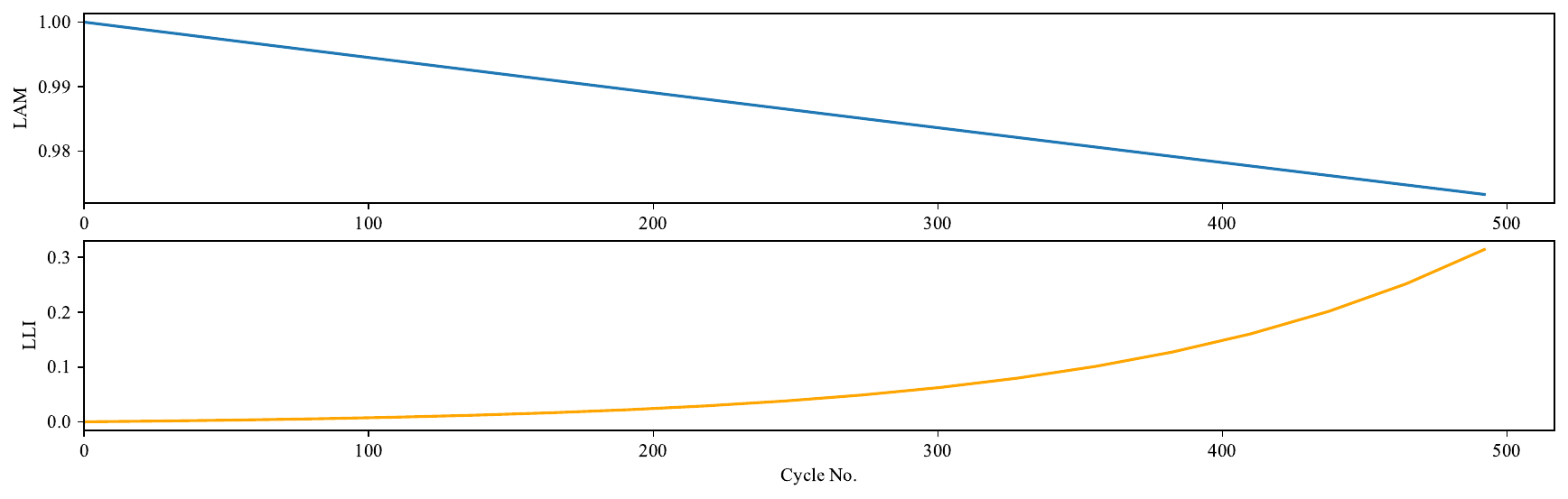}
    \caption{Quantitative results of degradation modes for capacity loss of Test LIB 1.}
    \label{fig:degradation_mode}
\end{figure}

Once the future degradation curve has been returned, the corresponding degradation modes that led to capacity fade can be deduced, as the parameters used to generate them are known. The results for Test LIB 1 can be seen in Fig.~\ref{fig:degradation_mode}, and values are given according to the quantification scheme in section \ref{sec_degradation_model}.

\subsection{Estimated Uncertainty}

\begin{figure}[ht]
    \centering
    \includegraphics[width=0.5\textwidth]{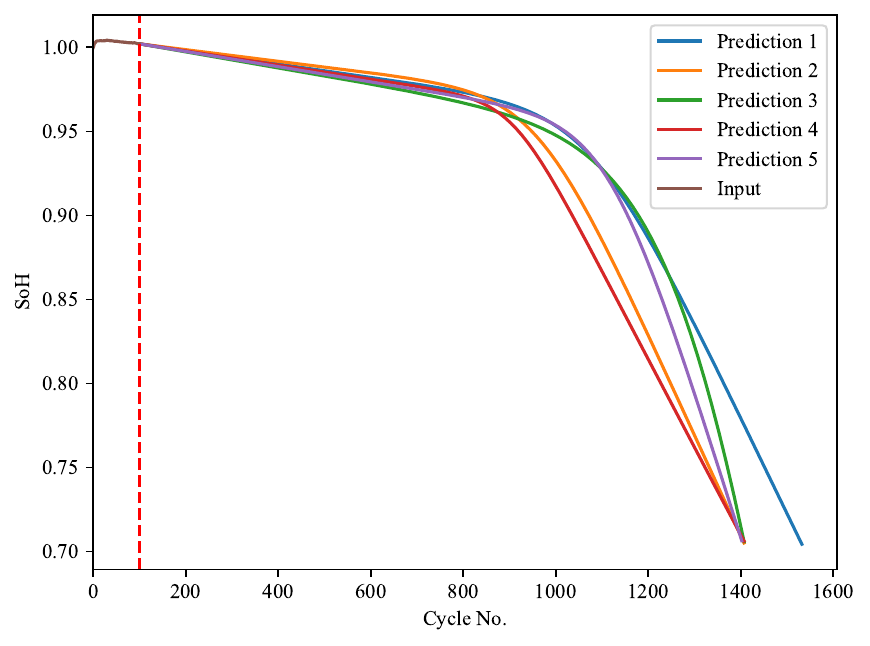}
    \caption{Top five most likely degradation paths for test LIB 2 where ``Prediction $k$'' represents the $k^{\text{th}}$ most likely scenario. They all have relatively similar shapes, indicating that the model has learned the degradation behaviour well.}
    \label{fig:uncertainty}
\end{figure}

\begin{figure}[ht]
    \centering
    \subfloat{
    \includegraphics[width=0.5\textwidth]{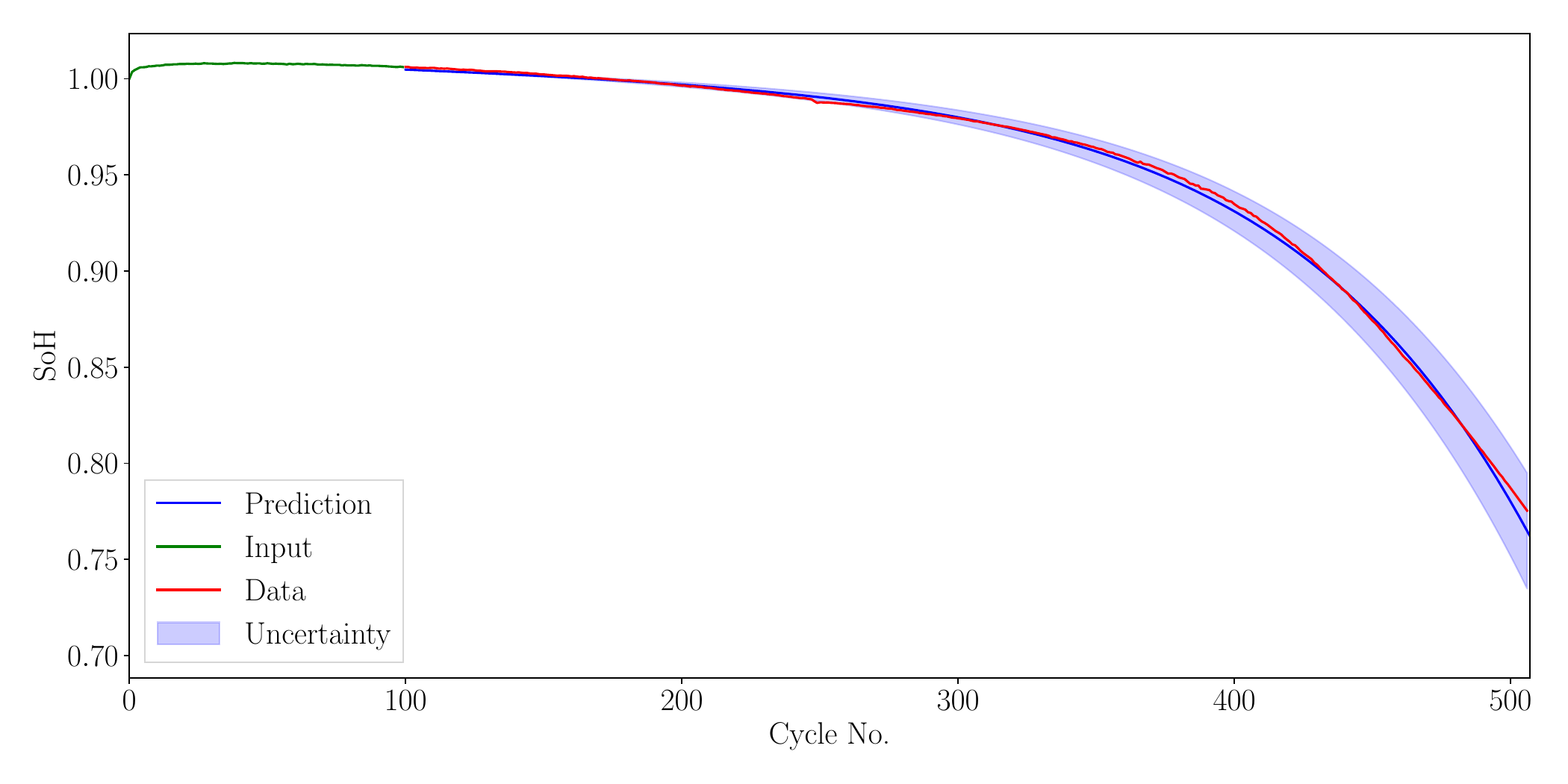}}
    \vskip -2pt
    \subfloat{
    \includegraphics[width=0.5\textwidth]{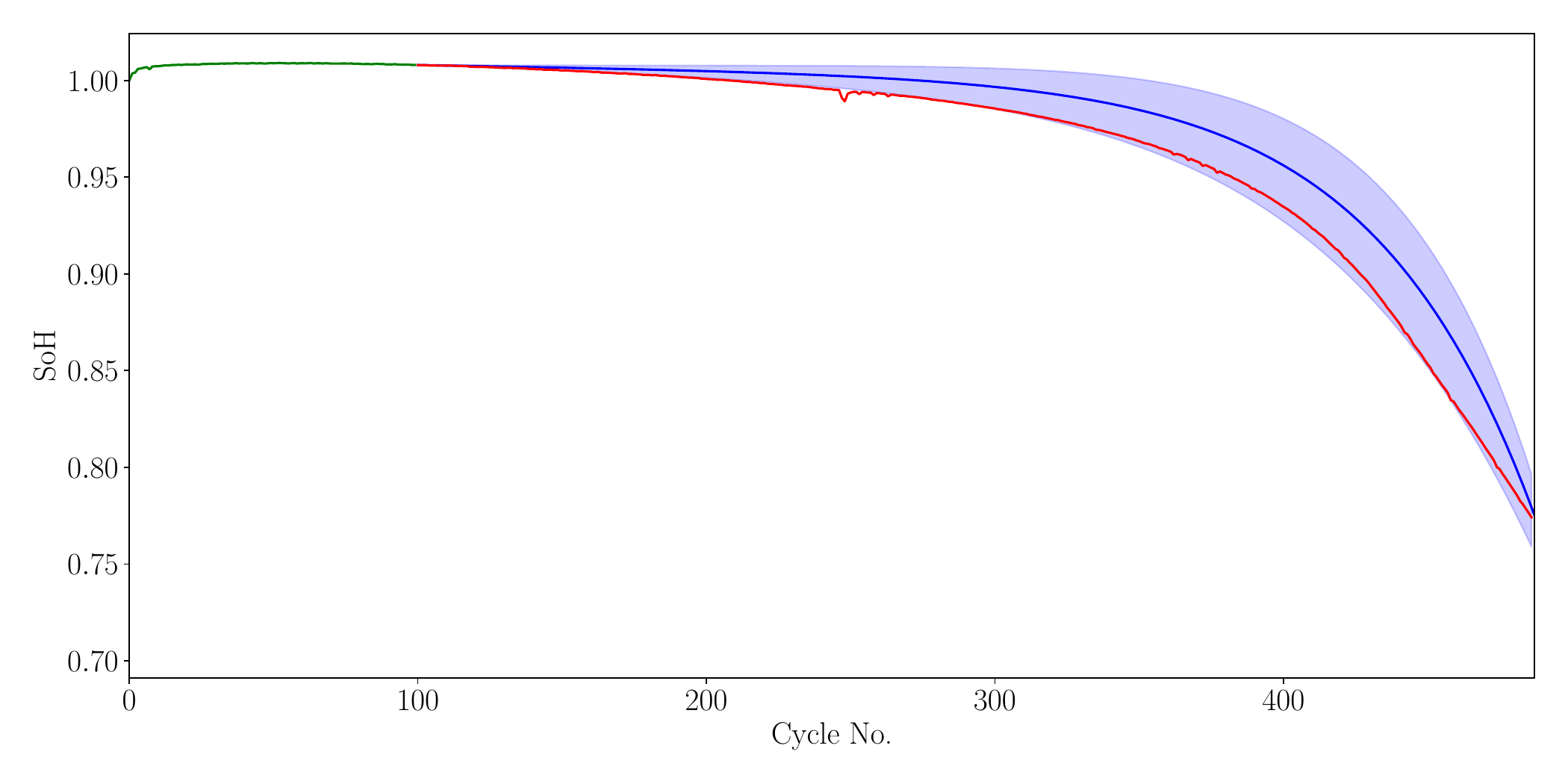}}
    \caption{Prediction for two test cells with calculated uncertainty.}
    \label{fig:uncertainty_calc}
\end{figure}

As the model returns a similarity score for each simulated scenario, it is possible to return a number of most likely degradation paths for each input. By returning more than one simulated scenario, it is possible to assess the consistency of the model's predictions. An example of the best five predictions for a single LIB is shown in Fig.~\ref{fig:uncertainty}. Furthermore, this process can be used to estimate how likely it is a particular LIB is to experience a knee-point.

Furthermore, when the model makes a prediction, additional degradation simulations can be generated by introducing small variations to the simulation parameters of the best-matching curve. These new curves are passed to the model and a new prediction is made. This process can be repeated many times, giving a distribution of selected curves from the many iterations of alternative simulations. This distribution reflects the prediction uncertainty resulting from the granularity of the simulations. Two examples are shown in Fig.~\ref{fig:uncertainty_calc}

\subsection{Zero-Shot} \label{sec:zero_shot}
Zero-shot estimation for LIB degradation is difficult due to inherent cross-domain variability between LIBs of different sizes and chemistries. Enabling zero-shot transfer to downstream tasks presents the possibility of reducing testing times in battery development cycles. To model the zero-shot capabilities of ACCEPT, we took two battery LIBs from \cite{attia2020closed}: one with a standard degradation profile and one with accelerated degradation. The purpose of this was to test whether our method could correctly differentiate between LIBs likely to experience the critical scenario of accelerated degradation. The corresponding simulated curves, one with accelerated degradation and one with standard degradation, were fed to the simulated curve encoder.

\begin{figure}[ht]
    \centering
    \includegraphics[width=0.5\textwidth]{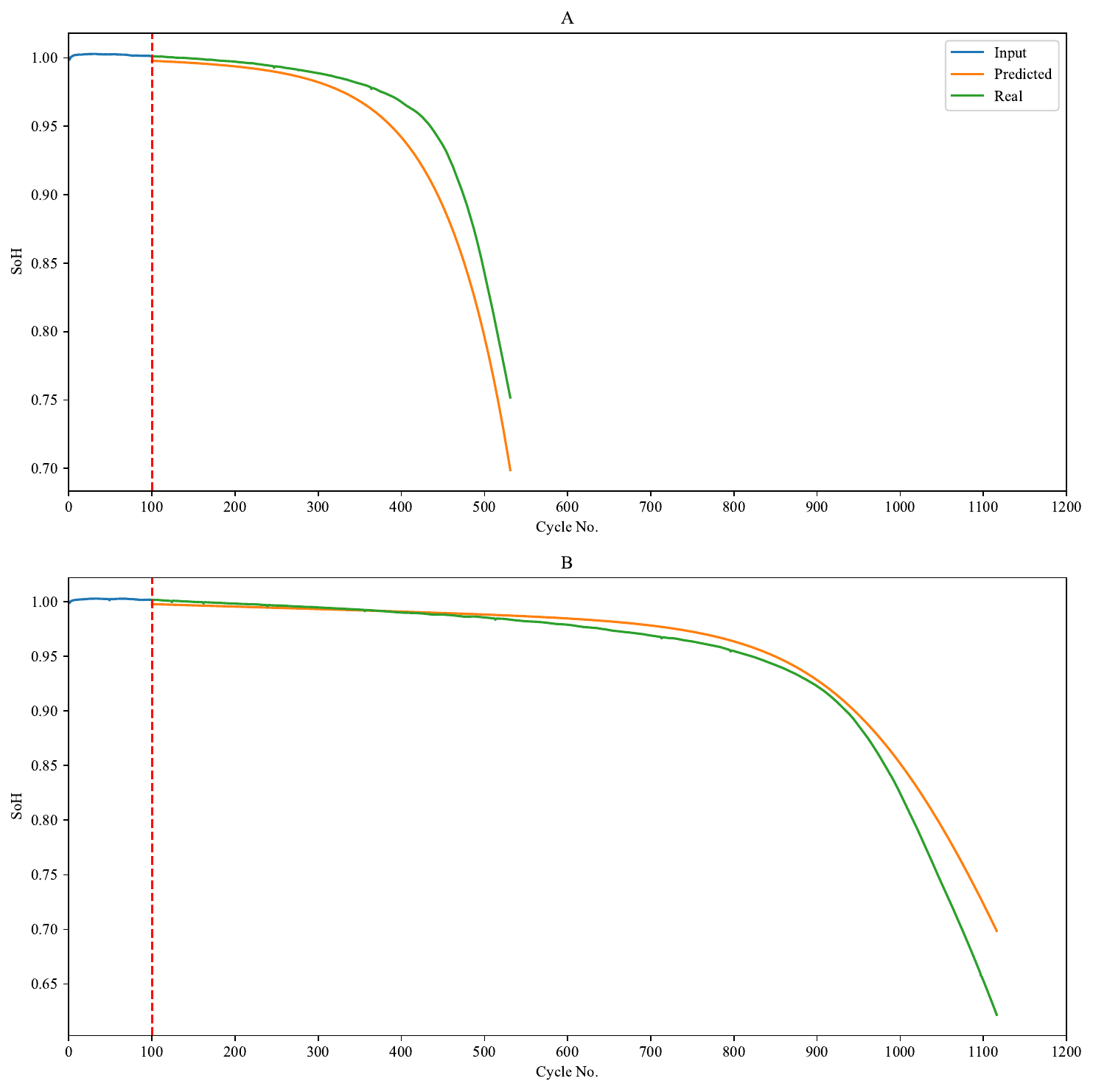}
    \caption{A) Zero-shot predictions against test LIB with accelerated degradation \& B) Zero-shot predictions against test LIB with standard degradation. Both LIBs taken from dataset from different source to the training set. The model was able to detect whether a particular LIB will experience accelerated degradation, a critical scenario to detect in the operation of LIBs}
    \label{fig:zeroshot}
\end{figure}

Despite the similarity in input between the two LIBs, ACCEPT was able to differentiate between the two profiles and chose the correct degradation pathway, as shown in Fig.~\ref{fig:zeroshot}. 

\section{Conclusion}
We have aimed to address the challenge of using data-driven techniques to accurately forecast degradation of Li-ion batteries and quantify the underlying electrochemical causes. Our method (ACCEPT) showed state-of-the-art accuracy for degradation modeling. Our approach differs from purely data-driven time-series by utilizing known physics to generate a number of simulated curves. ACCEPT then matched operational data to these simulations. As the underlying physical parameters of the model used to generate the simulated profiles are known, our framework made it possible to quantify the underlying degradation mode causing this. Our experiments showed that ACCEPT was able to generalize well to unseen scenarios, and could correctly anticipate if an LIB from an unseen dataset was likely to experience accelerated degradation, a critical phenomenon in the operation of EVs and BESSs. 
\subsection{Future Work}
This preliminary work aimed to demonstrate how a new class of machine learning models, paired with information from known physics about Li$^+$ batteries, can be used to accurately model degradation, whilst also quantifying the underlying degradation mode, something previous data-driven techniques have been unable to achieve. 
Currently, the model is purely trained on the Severson dataset; however the model can be fine-tuned on any dataset. Further open-source data on battery degradation could also be used to broaden the models applicability and increase its generalization to unseen LIB types. 
In this study, a TFT was used as the embedding model for the LIB's operational data. This model has the additional benefit of providing interpretable results through its variable selection network. This was not explored in this work, however future studies could use this to quantify the impact of different stress factors in the operational data on the overall degradation pathway. 
\\

\section{Impact statement}
This paper presents work whose goal is to advance application of machine learning to battery modeling. There are many potential societal consequences of this work. By improving the accuracy and efficiency of degradation forecasting, this work contributes to the development of more reliable and sustainable battery technologies, which are essential for reducing greenhouse gas emissions and enabling widespread adoption of clean energy solutions. We recognize the need for further research in this direction before practical use in production settings.

\newpage
\bibliographystyle{IEEEtran}
\bibliography{refs}

\end{document}